\pdfoutput=1
\documentclass{article}

\usepackage[dblblindworkshop,final]{neurips_2026}
\workshoptitle{AXIOM Foundations of Efficient Deep Learning}

\usepackage[utf8]{inputenc}
\usepackage[T1]{fontenc}
\usepackage{hyperref}
\usepackage{url}
\usepackage{booktabs}
\usepackage{amsmath}
\usepackage{graphicx}
\usepackage{subcaption}
\usepackage{flafter}
\usepackage{placeins}
\usepackage{microtype}
\usepackage{xcolor}
\definecolor{paperlink}{HTML}{1F4E79}
\hypersetup{
  colorlinks=true,
  linkcolor=paperlink,
  citecolor=paperlink,
  urlcolor=paperlink
}
\title{Memory-Efficient Speculative Decoding with Quantized Draft KV Cache}

\author{%
  Saksham Gupta\thanks{Equal contribution.} \\
  Carnegie Mellon University \\
  \texttt{sakshamg@andrew.cmu.edu} \\
  \And
  Michael Li\footnotemark[1] \\
  Carnegie Mellon University \\
  \texttt{ml7@andrew.cmu.edu} \\
}

\begin{document}

\maketitle

\begin{abstract}
Speculative decoding reduces generation cost, but it stores two KV caches: one
for the target model and one for the draft model. We reduce this extra memory by
quantizing the draft cache. The target cache and verifier remain in BF16, so that compression cannot change the target model's final
correction rule. We first study how
to divide precision between keys and values. A naive per-token quantizer makes
keys appear much more sensitive: splitting a 12-bit budget, K8V4 beats K4V8 in acceptance by 10.87~pp. However, with grouped
per-channel keys, the gap becomes $-0.25$~pp. A controlled Qwen test confirms
that changing only the key quantizer can erase or reverse the preferred bit
split. Hence, following KIVI, we propose K4V4 quantization with per-channel keys and per-token values. Across six model pairs, our method reduces draft cache memory by 66.20\%, and combined memory across both caches by 23.95\%. Mean acceptance drops from 57.93\% to 57.14\%, only 1.36\% relative. Overall, quantized draft caches can make speculative decoding more memory efficient; as a predictive principle for efficient AI design, key and value precision must be co-designed with the quantizer layout.
\end{abstract}

\section{Introduction}

Speculative Decoding (SD) improves LLM inference latency by letting a small \emph{draft} model propose tokens that a larger \emph{target} model verifies in parallel~\citep{leviathan2023speculative,
chen2023speculative}. Draft tokens are accepted sequentially until the first rejection, at which point a corrected token is sampled from the target model and generation continues from there. This maintains the target output rule while reducing the number of expensive target forward passes.

However, standard SD maintains a KV cache for each model, increasing memory pressure as context length and batch size grow. Let $\text{Mem}_T$ and $\text{Mem}_D$ be the target and draft cache sizes.
Standard SD stores $\text{Mem}_T+\text{Mem}_D$. We instead store $\text{Mem}_T+\widehat \text{Mem}_D$,
where $\widehat \text{Mem}_D$ is the low-precision draft cache size. Note that the target cache remains unchanged. Quantization may reduce draft model acceptance, but it will not affect target model performance.

The main design question is how to quantize keys and values. Attention computes $O=AVW_O$ with $A=\operatorname{softmax}(QK^\top/\sqrt{d_h}+M)$~\citep{vaswani2023attention}. Keys control where we read, while values contain the information returned by reads. Empirically, it is tempting to conclude that keys
always need more precision because they are more sensitive to noise. For example, under Gaussian noise with standard deviation 0.03 times the per-layer RMS, key noise reduced Qwen2.5-1.5B's top-1 agreement with the noise-free cache to 61.7\%, while value noise retained 99.2\% (Appendix~\ref{app:protocol}).

We organize our study around three research questions. \textbf{(1) Key and value precision.} Is the apparent need for higher key precision intrinsic to attention, or is it caused by quantizer geometry and outlier handling? \textbf{(2) Memory and acceptance.} How much SD memory can draft-only KV cache quantization save without materially reducing token acceptance? \textbf{(3) Generalization and practical savings.} Do these memory-quality tradeoffs hold across model families, long contexts, and retrieval workloads, and do they translate into real packed-storage savings?

\section{Related Work}

In standard autoregressive inference, KV cache quantization successfully mitigates memory costs, though precision allocation is often heuristically driven. For example, KIVI applies an asymmetric strategy of per-channel quantization for keys and per-token quantization for values \citep{liu2024kivi}, whereas Quantize What Counts argues for allocating higher precision to keys based on the distinct geometries of key and value representations \citep{hariri2026quantize}. Other approaches optimize bit allocation dynamically: KVTuner searches for layer-wise precision pairings \citep{li2025kvtuner}, RateQuant distributes bits across attention heads via rate-distortion theory \citep{zuo2026ratequant}, and Block-GTQ allocates precision across RoPE frequency blocks \citep{liang2026blockgtq}. We build on KIVI's geometry and study asymmetric bit splitting (e.g., K8V4 vs. K4V8) in conjunction with it.

While QuantSpec introduces hierarchical KV quantization for self-speculative decoding \citep{tiwari2025quantspec}, our work addresses a different paradigm. We study standard two-model speculative decoding, specifically targeting the quantization of the draft model's cache. By evaluating the trade-off between total memory savings and candidate acceptance rates, we isolate how key-quantizer geometry alters conclusions about the optimal precision balance between keys and values.

\section{Draft Cache Quantization}

\begin{figure}[ht]
  \centering
  \includegraphics[width=0.8\linewidth,trim=0 125 0 140,clip]{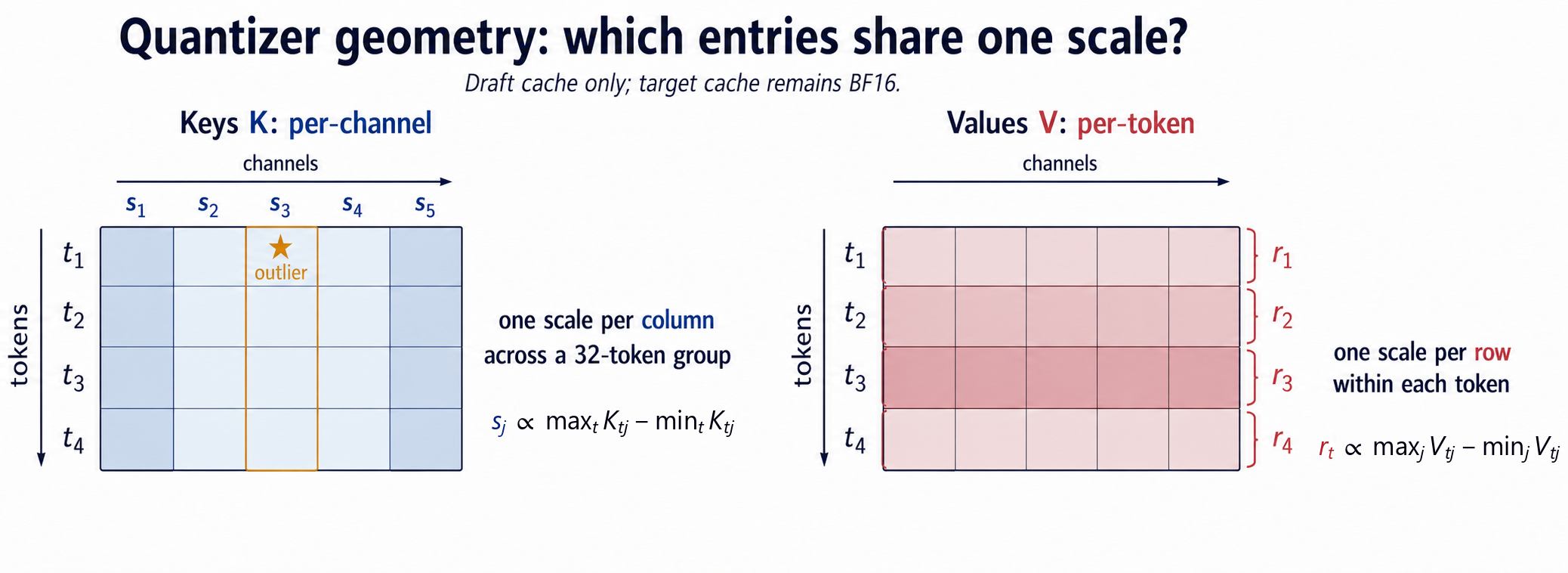}
  \caption{\textbf{Quantizer geometry.} Draft keys share scales by channel,
  while draft values share scales by token. K8V4 and K4V8 allocate equal
  average payload precision.}
  \label{fig:quantizer-geometry}
\end{figure}

Let K$a$V$b$ denote $a$-bit keys and $b$-bit values. We first compare K8V4 against K4V8 to test whether keys or values need more bits. Our simple baseline uses one symmetric scale for each token and KV head. For a
$b$-bit group $G$, symmetric scale $s_G$ is computed as follows:
\begin{equation}
 s_G=\frac{\max_{x\in G}|x|}{2^{b-1}-1},\qquad
 \widehat x=s_G\operatorname{clip}(\operatorname{round}(x/s_G)).
 \label{eq:symmetric}
\end{equation}
This baseline lets channel-specific key outliers set the scale for the full
vector. Inspired by KIVI, our layout instead uses 32-token per-channel key groups
and per-token value groups within each head~\citep{liu2024kivi}, as
shown in Figure~\ref{fig:quantizer-geometry}. Both are affine: groups store a zero-point $m_G=\min_{x\in G}x$ and a scale $s_G=(\max_{x\in G}x-m_G)/(2^{b}-1)$ (Appendix~\ref{app:protocol}). The 128 most recent keys stay in a BF16 residual window. Memory results count payload bits, BF16 scales, zero-points, and this window.

The target cache and verifier remain in BF16. We measure acceptance (the fraction of proposed draft tokens that the target accepts),
KL from BF16 draft logits, top-1 agreement, and retrieval accuracy.

\section{Evaluation}

We test six target and draft pairs across Qwen2.5~\citep{yang2024qwen25},
Qwen3~\citep{yang2025qwen3}, Llama 3~\citep{grattafiori2024llama3}, OLMo
2~\citep{olmo2025furious}, and SmolLM2~\citep{allal2025smollm2}. Model sizes
range from 360M to 8B parameters. We evaluate 576 FineWeb-Edu prompts of 1,024 tokens (96 per
pair)~\citep{penedo2024fineweb} with greedy decoding. The draft proposes up to four
tokens per round, and each run generates 16 tokens. Every prompt is tested with BF16,
K8V4, K4V8, and K4V4. The noise test, the RQ1 acceptance comparisons, and the
KL and top-1 measurements use WikiText-2 validation
text~\citep{merity2016pointer}. The equal-model macro is a macro average that weights model pairs equally.

With the Qwen2.5 3B/1.5B pair, we also run a controlled test that changes only the key quantizer, 192 paired passkey retrieval examples at 16K context, 256-token continuations, and an end-to-end packed prototype (Appendix~\ref{app:followup}). Our sequential verifier makes one target call per emitted token, so it measures draft quality but not target-call savings. Our batched verifier scores each proposal in one target call and rolls back the target cache; only the RQ1 acceptance comparisons, the prototype, and the Spearman grid of Appendix~\ref{app:objectives} use it.

\section{Results}
\label{sec:results}

\subsection{RQ1: Are keys intrinsically more precision-sensitive?}

\begin{figure}[!ht]
  \centering
  \includegraphics[width=\linewidth]{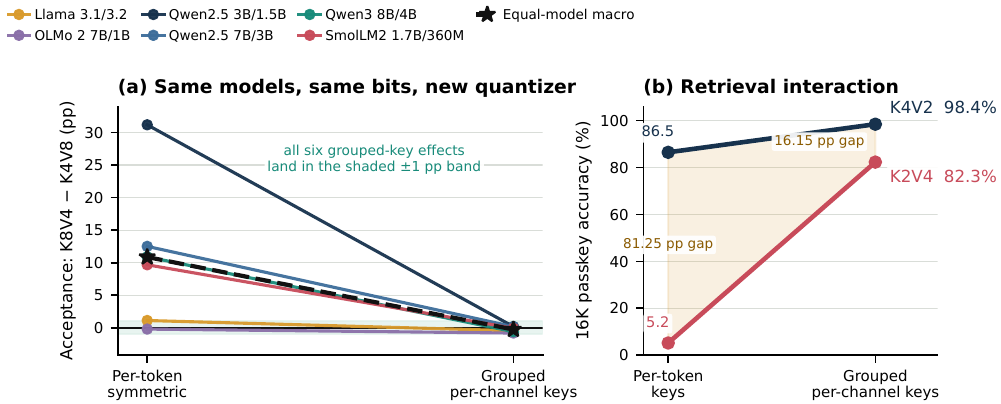}
  \caption{\textbf{Quantizer geometry changes the K/V conclusion.} (a) Each
  line follows one model pair while models and bit budgets remain fixed.
  Positive values favor more key bits. (b) On 192 paired Qwen2.5 passkey
  examples, grouped per-channel keys recover most of the failure of K2V4 and
  sharply reduce the apparent key-bit advantage.}
  \label{fig:geometry}
\end{figure}

We demonstrate that the apparent need for higher key precision depends strongly on quantizer geometry, and is not necessarily an intrinsic property of attention. When using a naive per-token symmetric quantizer, keys appear highly sensitive: K8V4 outperforms an equal-mean-bit K4V8 by 10.87~pp on average. However, switching to grouped per-channel keys eliminates this advantage, shifting the contrast to $-0.25$~pp and dropping all within-pair gaps below 1~pp (Figure~\ref{fig:geometry}a).

A controlled Qwen intervention, which strictly isolates the key quantizer, confirms this behavior. Figure~\ref{fig:geometry}a's settings also differ in value quantizer, residual window, and prompts, but here both use affine values, no residual window, and the same prompts. Changing the key geometry shifts the K8V4 versus K4V8 gap from $+28.70$ ($+31.18$ in Figure~\ref{fig:geometry}a) to $-0.33$~pp, and reverses the aggressive K4V2 versus K2V4 contrast from $+15.05$ to $-3.66$~pp. Both settings share the batched verifier, and the interaction extends to verifier-free 16K retrieval (Figure~\ref{fig:geometry}b), where modifying key geometry shrinks the K4V2 versus K2V4 accuracy gap from 81.25 to 16.15~pp. Ultimately, bit width and quantizer geometry must be co-designed; keys do not universally demand more precision (Appendix~\ref{app:geometry}).

\FloatBarrier

\subsection{RQ2: How much memory can draft-only quantization save?}

Using K4V4 as our balanced operating point, we find it removes about two-thirds of draft cache memory. Across all six evaluated model pairs, K4V4 reduces draft cache bytes by 66.20\% and the combined target-plus-draft KV memory by 23.95\% (11.30\% to 33.17\% across pairs).

\begin{figure}[!ht]
  \centering
  \includegraphics[width=\linewidth]{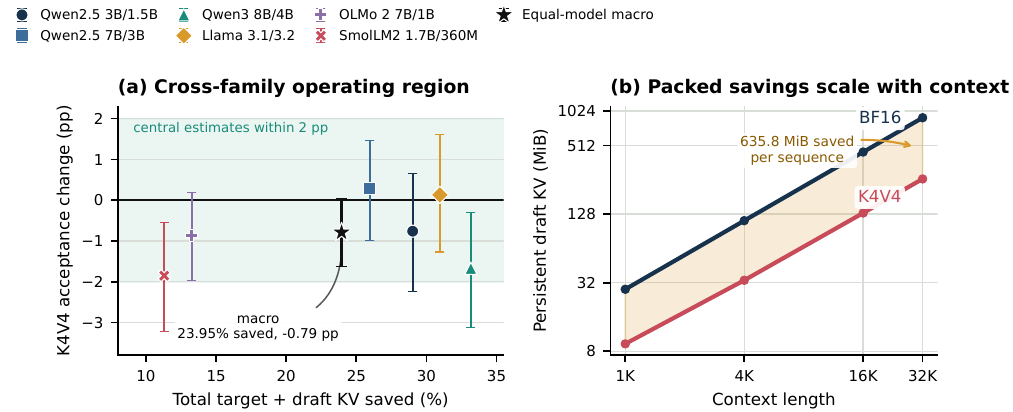}
  \caption{\textbf{Memory and acceptance.} (a) K4V4 total KV saving and paired
  acceptance change across six target/draft pairs. Error bars are paired 95\%
  CIs; the star is the equal-model macro. (b) Actual bit-packed draft cache
  storage, including quantizer metadata and the BF16 residual window. The
  absolute saving grows with retained context.}
  \label{fig:memory-acceptance}
\end{figure}

Importantly, this compression incurs minimal degradation to candidate generation. The mean cross-family acceptance rate falls only from 57.93\% to 57.14\%, a 1.36\% relative drop (Figure~\ref{fig:memory-acceptance}a). In separate WikiText-2 runs, the K4V4 draft's KL from BF16 is 0.00730 (95\% CI $[0.00660,0.00801]$), with 95.99\% top-1 agreement. With 256-token continuations on one Qwen2.5 pair, the acceptance change ranges from $-0.59$ to $+0.33$~pp, and the outputs match target-only greedy decoding under the sequential verifier (Appendix~\ref{app:followup}).

\FloatBarrier

\subsection{RQ3: Does the result generalize and produce real savings?}

The memory-quality tradeoffs of K4V4 hold across six model pairs and extended context windows. At 16K and 32K context with the Qwen2.5 3B/1.5B pair, K4V4 reduces combined KV memory by about 31\%, with mean acceptance drops of only 0.94 and 0.73~pp, respectively (Appendices~\ref{app:geometry} and~\ref{app:retrieval}).

The analytical byte savings also hold in physical packed storage. At 32K context, a standard Qwen2.5-1.5B BF16 cache requires 896.0 MiB per sequence, whereas our packed K4V4 implementation consumes just 260.2 MiB, saving 635.8 MiB (Figure~\ref{fig:memory-acceptance}b). However, these savings do not yet translate into a serving benefit. In an end-to-end prototype with the same pair at 1K and 16K context, packed K4V4 stores 67\% to 71\% fewer draft cache bytes, but it is 18\% to 33\% slower than BF16 speculative decoding, which is slower than target-only decoding. Peak GPU memory is also unchanged, since the draft cache is packed only after a BF16 prefill. Its batched verifier reproduces target-only outputs on only 2 to 5 prompts per setting (of 12 or 24), even with a BF16 draft, so we treat these timings as preliminary (Appendix~\ref{app:followup}). Appendix~\ref{app:objectives} covers policy selection.

\section{Discussion}

\textbf{Conclusion.} Draft-only KV cache quantization can substantially reduce
the memory overhead of two-model speculative decoding while largely preserving draft
acceptance. K4V4 removes about two-thirds of draft cache memory, while
the target verifier and its correction rule remain unchanged. The results also
show that key and value precision cannot be evaluated independently of
quantizer geometry,
providing a predictive principle for designing future quantization systems.

\textbf{Limitations.} Our longest-context experiments have fewer prompts than the 1K study. The
retrieval task is partly synthetic. Fake quantization, which stores dequantized values in BF16, measures quality but not
production speed. Our packed prototype is still slower than BF16
speculative decoding and does not yet lower peak GPU memory. The 256-token and prototype experiments cover one model pair at batch size one, and the batched verifier used for RQ1 is not yet exact. We do not measure key outliers directly, and keys and values may differ intrinsically in settings we do not test.

\textbf{Future work.} Future work could replace static bit assignments with serving-time adaptive
computation, building predictive models for runtime quantization that select key and value precision by layer, context region, or
request. This will also require an exact batched verifier, streaming prefill, fused packed-cache attention kernels, and end-to-end evaluation
in production systems.

\clearpage
\bibliography{references}
\bibliographystyle{plainnat}

\clearpage
\appendix

\section{Experimental Details}
\label{app:protocol}

\paragraph{Model pairs and data.}
Table~\ref{tab:app-models} lists the six model pairs. Only the draft cache is
quantized. The target model verifies tokens with its original BF16 cache. Each
run prefills a 1,024-token FineWeb-Edu prompt, proposes up to four tokens per
round, and generates 16 tokens. Across six model pairs, we evaluate 576 prompts.
Every prompt is tested with BF16, K8V4, K4V8, and K4V4.
Prompts are consecutive 1,024-token blocks from the streamed
\texttt{sample-10BT} training split. We tokenize with the target tokenizer
without special tokens and place an end-of-sequence token between documents.
Each pair uses three disjoint shards of 32 prompts at fixed stream offsets. The
noise test, Tables~\ref{tab:app-geometry} and~\ref{tab:app-factorial}, the KL and
top-1 measurements, and the Spearman grid use WikiText-2 validation text.
These prompts have 1,024 tokens, except in the noise test (256) and the 4K
setting of the Spearman grid. The noise test and the per-token column of
Table~\ref{tab:app-geometry} take consecutive blocks; the other WikiText-2
experiments shuffle blocks with a per-run seed. The
Qwen3 checkpoints are instruction-tuned, and all others are base models. All
models read raw text without chat templates.

\begin{table}[ht]
  \centering
  \small
  \caption{Primary speculative-decoding model pairs (Hugging Face checkpoint
  identifiers).}
  \label{tab:app-models}
  \begin{tabular}{lll}
    \toprule
    Family & Target & Draft \\
    \midrule
    Qwen2.5 & \texttt{Qwen/Qwen2.5-3B} & \texttt{Qwen/Qwen2.5-1.5B} \\
    Qwen2.5 & \texttt{Qwen/Qwen2.5-7B} & \texttt{Qwen/Qwen2.5-3B} \\
    Qwen3 & \texttt{Qwen/Qwen3-8B} & \texttt{Qwen/Qwen3-4B} \\
    Llama 3 & \texttt{meta-llama/Llama-3.1-8B} & \texttt{meta-llama/Llama-3.2-3B} \\
    OLMo 2 & \texttt{allenai/OLMo-2-1124-7B} & \texttt{allenai/OLMo-2-0425-1B} \\
    SmolLM2 & \texttt{HuggingFaceTB/SmolLM2-1.7B} & \texttt{HuggingFaceTB/SmolLM2-360M} \\
    \bottomrule
  \end{tabular}
\end{table}

\paragraph{Decoding protocol.}
All speculative decoding experiments use greedy decoding at batch size one. In
each round, the draft proposes up to four argmax tokens. The target accepts the
longest proposal prefix that matches its own argmax tokens. The target
then emits a correction token at the first disagreement, or a bonus token after
a fully accepted proposal. We crop the draft cache to the committed prefix, which
discards rejected entries, and append the emitted token to both caches.
Generation has a fixed length without early stopping. Acceptance excludes
correction and bonus tokens. The sequential verifier advances the target one
accepted token at a time and makes one target call per emitted token, so it
measures draft quality but not target-call savings. The batched verifier scores
a proposal in one target call and rolls the target cache back. Tables~\ref{tab:app-geometry} and~\ref{tab:app-factorial}, the Spearman grid,
and the prototype use the batched verifier; all other speculative results use
the sequential verifier.

\paragraph{Quantizer geometries.}
The simple quantizer (Equation~\ref{eq:symmetric}) uses one scale for each
token and KV head, for keys and values. The stronger
quantizer uses KIVI's per-channel key and per-token value geometry
\citep{liu2024kivi} with affine groups (Equation~\ref{eq:affine}). In our implementation, keys are grouped over 32 tokens,
values are grouped within each token and head, and the newest 128 key tokens
remain in BF16. In the controlled test, we remove this residual window and
change only the key quantizer (per-token symmetric versus per-channel
affine). We never update model weights.

\paragraph{Quantizer specification.}
An affine $b$-bit group $G$ has zero-point $m_G=\min_{x\in G}x$ and scale $s_G$:
\begin{equation}
 s_G=\frac{\max_{x\in G}x-m_G}{2^{b}-1},\qquad
 \widehat x=m_G+s_G\operatorname{clip}\big(\operatorname{round}((x-m_G)/s_G),\,0,\,2^{b}-1\big).
 \label{eq:affine}
\end{equation}
Consider one layer with keys and values of shape $H\times L\times d$, with $H$
KV heads, $L$ committed tokens, and $d$ channels. We use K$a$V$b$ with group
size $g=32$ and residual length $R=128$. For keys, we set
$P=g\lfloor\max(0,L-R)/g\rfloor$. A key group covers one head, one channel, and
$g$ consecutive tokens. We apply Equation~\ref{eq:affine} with $a$ bits to each
newly completed group below $P$. Tokens at or beyond $P$ remain in BF16, and we
never pad or requantize incomplete groups. For values, we apply
Equation~\ref{eq:affine} with $b$ bits to the $d$ channels of each new token and
head. Values have no residual window. During a speculative round, $P$ is
frozen, so we only quantize committed keys. We compute statistics in FP32 and
floor ranges at $10^{-8}$. We round to nearest with ties to even. The
symmetric baseline clips to $\pm(2^{b-1}-1)$. Quality experiments use fake
quantization, which writes dequantized tensors back into the BF16 cache and
computes bytes analytically. The packed implementation stores two 4-bit payload
entries per byte in row-major \texttt{uint8} tensors. Metadata and residual
keys remain in BF16.

\paragraph{Byte accounting.}
We count packed payloads, BF16 scales, zero-points, and the BF16 residual window.
Let $\text{Mem}_T$ be the target cache, $\text{Mem}_D$ be the native draft cache, and
$\widehat \text{Mem}_D$ be the quantized draft cache. Total memory saving is
\begin{equation}
  \operatorname{saved}_{\mathrm{total}}
  =1-\frac{\text{Mem}_T+\widehat \text{Mem}_D}{\text{Mem}_T+\text{Mem}_D}.
\end{equation}
The draft cache shrinks by 66\% to 71\%, depending on experimental factors. The target cache does not shrink, so the combined saving across both caches is smaller.
For each pair, the total saving equals the draft saving times the draft's share
of the combined BF16 cache (Table~\ref{tab:app-k4v4-generalization}). All
savings refer to KV cache bytes, not model weights or activations.

\paragraph{Metrics and uncertainty.}
Acceptance is the fraction of proposed draft tokens accepted by the target.
Acceptance differences are absolute, in percentage points (pp), unless marked as relative.
We also measure KL from the BF16 draft, top-1 agreement, and exact retrieval
accuracy. KL is $\mathrm{KL}(p\,\|\,q)$, where $p$ and $q$ are the draft's
next-token distributions with a BF16 and a quantized cache on 32-token
teacher-forced continuations. We average it over positions and prompts, and then
equally over model pairs. Top-1 agreement is the fraction of these positions with
the same argmax.
Confidence intervals are paired bootstrap intervals over prompts with 5,000 to
30,000 resamples. An equal-model macro averages a metric over model pairs with
equal weight. The macro of Table~\ref{tab:app-k4v4-generalization} resamples
pairs and then prompts; other macros resample pairs only.

\paragraph{Noise test.}
The introduction's noise test uses Qwen2.5-1.5B, the draft of our first pair, on
the first 128 consecutive 256-token blocks of WikiText-2 validation text. For
each block, we prefill 255 tokens, add Gaussian noise to the cached keys, the
cached values, or both in every layer, and compare the next-token distribution
with that of the noise-free cache. The noise has standard deviation $\alpha$
times the RMS of that layer's cached keys or values. Top-1 agreement is the
fraction of blocks whose argmax is unchanged. Up to $\alpha=0.01$, key and value
noise both keep at least 97.7\% top-1 agreement. At $\alpha=0.03$, key noise
keeps 61.7\% (KL 1.46) and value noise keeps 99.2\% (KL 0.0016); noise on both
keeps 62.5\%. At $\alpha=0.1$, value noise still keeps 95.3\%, while key noise
keeps 10.9\%.

\paragraph{Verifier audit.}
The target still checks every proposed token, so draft quantization does not
change the mathematical correction rule. We compare every speculative output
with an independent tokenwise target path and call a mismatch a near tie when
the target's top-1 margin is at most $10^{-3}$. All sequential-verifier
trajectories match, including 38,400 checks in Appendix~\ref{app:objectives}.
With the batched verifier, Table~\ref{tab:app-factorial} and the grouped-key
column of Table~\ref{tab:app-geometry} drop prompts with non-tie mismatches;
Table~\ref{tab:app-geometry}'s per-token column keeps them (one to three prompts
in four of its twelve runs). A BF16 target
cache preserves the correction rule, but it does not guarantee bit-identical
outputs from a batched verifier. Within each RQ1 comparison, both quantizer arms
use the batched verifier, and the passkey test, which needs no verifier, shows
the same key-quantizer interaction (Appendix~\ref{app:retrieval}).

\paragraph{Reproducibility details.}
We release the
implementation, SLURM launchers, pinned environment, tests, aggregation
code, and output schemas at \href{https://github.com/sgdescent/KV_Reduce}{github.com/sgdescent/KV\_Reduce} under the MIT License.
The released launcher for Table~\ref{tab:app-factorial} and the Spearman grid
now uses consecutive shards and the sequential verifier; the reported runs
predate this change.

\section{Cross-Family Generalization and Mechanism Ablations}
\label{app:geometry}

The cross-family study tests
whether the K4V4 operating point and the K/V contrast generalize beyond one
Qwen pair. The controlled Qwen intervention then isolates mechanism by changing
only the key quantizer. Group-size and residual-window sweeps are implementation
ablations rather than additional generalization claims.

\paragraph{Cross-family generalization test.}
Table~\ref{tab:app-k4v4-generalization} reports the K4V4 operating point for
all six speculative model pairs. Four pair-level confidence intervals include zero;
Qwen3 and SmolLM2 show small resolved losses. All six central estimates remain
within 2~pp, while combined savings vary with the relative
size of the target cache and draft cache, from 33.17\% at a 50.0\%
draft share to 11.30\% at a 17.2\% draft share.

\begin{table}[ht]
  \centering
  \small
  \caption{Cross-family K4V4 generalization. Acceptance changes are differences in pp
  from the BF16 draft cache. BF16 and K4V4 are acceptance rates in \%.}
  \label{tab:app-k4v4-generalization}
  \resizebox{\linewidth}{!}{%
  \begin{tabular}{lrrrrrr}
    \toprule
    Target / draft & BF16 & K4V4 & $\Delta$ acceptance (95\% CI) & Draft saved & Draft share & Total saved \\
    \midrule
    Qwen2.5 3B/1.5B & 59.78 & 59.02 & $-0.76\ [-2.23,+0.66]$ & 66.33\% & 43.8\% & 29.02\% \\
    Qwen2.5 7B/3B & 61.80 & 62.08 & $+0.28\ [-0.98,+1.47]$ & 66.33\% & 39.1\% & 25.96\% \\
    Qwen3 8B/4B & 54.06 & 52.39 & $-1.67\ [-3.12,-0.30]$ & 66.33\% & 50.0\% & 33.17\% \\
    Llama 3.1/3.2 & 64.95 & 65.09 & $+0.13\ [-1.27,+1.60]$ & 66.33\% & 46.7\% & 30.96\% \\
    OLMo 2 7B/1B & 54.03 & 53.17 & $-0.86\ [-1.98,+0.18]$ & 66.33\% & 20.0\% & 13.27\% \\
    SmolLM2 1.7B/360M & 52.95 & 51.10 & $-1.85\ [-3.21,-0.54]$ & 65.55\% & 17.2\% & 11.30\% \\
    \midrule
    Equal-model macro & 57.93 & 57.14 & $-0.79\ [-1.63,+0.04]$ & 66.20\% & 36.1\% & 23.95\% \\
    \bottomrule
  \end{tabular}}
\end{table}
\FloatBarrier

\paragraph{Cross-family quantizer-geometry test.}
Table~\ref{tab:app-geometry} gives every equal-bit K8V4 versus K4V8 result.
The per-token quantizer makes extra key bits look valuable for Qwen2.5, Qwen3, and
SmolLM2. With grouped per-channel keys, every difference is below 1~pp.
Only the models, bit budgets, and verifier match: the grouped-key column also
uses affine values, a 128-token BF16 residual window, and 64 shuffled instead of
32 consecutive blocks per pair. Table~\ref{tab:app-factorial} isolates the key
quantizer.

\begin{table}[ht]
  \centering
  \small
  \caption{Acceptance contrasts in pp. ``Shift'' is grouped
  per-channel minus per-token. Brackets give paired 95\% CIs for each
  within-pair contrast.}
  \label{tab:app-geometry}
  \resizebox{\linewidth}{!}{%
  \begin{tabular}{lrrr}
    \toprule
    Target / draft & Per-token $\Delta$ & Grouped-key $\Delta$ & Shift \\
    \midrule
    Qwen2.5 3B/1.5B & $+31.18\ [23.33,39.00]$ & $+0.20\ [-1.28,1.68]$ & $-30.99$ \\
    Qwen2.5 7B/3B & $+12.51\ [7.54,17.84]$ & $+0.18\ [-1.09,1.50]$ & $-12.33$ \\
    Qwen3 8B/4B & $+10.91\ [4.61,17.27]$ & $-0.58\ [-1.88,0.64]$ & $-11.49$ \\
    Llama 3.1/3.2 & $+1.13\ [-1.91,4.28]$ & $-0.44\ [-1.94,1.06]$ & $-1.57$ \\
    OLMo 2 7B/1B & $-0.19\ [-2.30,2.05]$ & $-0.80\ [-2.22,0.45]$ & $-0.61$ \\
    SmolLM2 1.7B/360M & $+9.69\ [5.36,14.31]$ & $-0.06\ [-0.98,0.84]$ & $-9.75$ \\
    \midrule
    Equal-model macro & $+10.87\ [3.75,19.68]$ & $-0.25\ [-0.56,0.04]$ & $-11.12\ [-19.71,-4.25]$ \\
    \bottomrule
  \end{tabular}}
\end{table}

\paragraph{Controlled Qwen test.}
Table~\ref{tab:app-factorial} keeps the models, prompts, value quantizer, bit
budget, and decoding loop fixed. Only the key quantizer changes. Both arms use
affine per-token values, no residual window, and three seeds of 64 shuffled
prompts. The per-token K8V4 versus K4V8 gap of $+28.70$~pp therefore differs
from the $+31.18$~pp in Table~\ref{tab:app-geometry}, which uses symmetric
values and other prompts. The K4V2
versus K2V4 result changes sign, so the preferred bit split also changes.

\begin{table}[ht]
  \centering
  \small
  \caption{Qwen2.5 3B/1.5B key-axis intervention. Values are acceptance
  contrasts in pp with paired 95\% CIs.}
  \label{tab:app-factorial}
  \begin{tabular}{lrr}
    \toprule
    Contrast & Per-token keys & Grouped per-channel keys \\
    \midrule
    K8V4 $-$ K4V8 & $+28.70\ [26.10,31.38]$ & $-0.33\ [-1.26,0.62]$ \\
    K4V2 $-$ K2V4 & $+15.05\ [13.58,16.65]$ & $-3.66\ [-5.80,-1.56]$ \\
    \bottomrule
  \end{tabular}
\end{table}

\paragraph{Granularity and residual-window controls.}
With the Qwen2.5 3B/1.5B pair on FineWeb-Edu, we also test key groups of 16, 32, 64, and 128 tokens. The K4V2
versus K2V4 differences are $-5.83$, $-2.94$, $-1.57$, and $-1.87$~pp; only
the 16-token interval excludes zero. We test BF16 residual windows of 0, 32, 128, and
256 tokens as well. A longer window gives no clear K4V4 acceptance gain.
Removing it increases draft cache saving from 66.33\% to 70.56\%.
Our 256-token experiments also show no clear acceptance gain from the window
(Appendix~\ref{app:followup}), so we treat 128 tokens as a configuration
choice rather than an optimum.

\section{Long-Context Generalization and Retrieval Stress Tests}
\label{app:retrieval}

\paragraph{16K associative retrieval.}
Table~\ref{tab:app-passkey} reports 192 examples per model. Each example is a
16,384-token prompt. The prompt hides 16 records of the form ``Archive record
[tag]: the access code is [six digits]'' at even spacing in filler text and
then asks for the code of one tagged record. The 15 distractor codes differ
from the correct code in one digit. We query the 3rd, 9th, or 15th record.
We count an example as retrieved when the correct code has the highest
length-normalized log-likelihood among the 16 candidates. We run each draft
model alone on three seeds of 64 examples. Qwen2.5 benefits
from giving keys more bits. Llama and Qwen3 do not show the same effect. On
Qwen2.5, per-token K4V2 and K2V4 retrieve 166 and 10 examples. Grouped-key K4V2
and K2V4 retrieve 189 and 158. The key quantizer changes the gap by 65.10~pp
(95\% CI $[56.77,73.44]$).

\begin{table}[ht]
  \centering
  \small
  \caption{Associative-passkey accuracy at 16K with grouped per-channel keys.}
  \label{tab:app-passkey}
  \begin{tabular}{lrrrr}
    \toprule
    Draft model & BF16 & K4V2 & K2V4 & $\Delta$ (pp) \\
    \midrule
    Qwen2.5-1.5B & 186/192 & 189/192 & 158/192 & $+16.15$ \\
    Llama-3.2-3B & n/a & 185/192 & 183/192 & $+1.04$ \\
    Qwen3-4B & 191/192 & 192/192 & 192/192 & $0.00$ \\
    \bottomrule
  \end{tabular}
\end{table}

We also run a natural LongBench passage-retrieval task~\citep{bai2024longbench}
on three 7B to 8B instruction models (Qwen2.5-7B-Instruct,
Llama-3.1-8B-Instruct, and Qwen3-8B). The task is
\texttt{passage\_retrieval\_en}, which asks for the source paragraph of an
abstract among 30 Wikipedia paragraphs. We use the LongBench prompt template
without a chat template and truncate prompts in the middle to at most 32,752
tokens. We score the 30 answers by length-normalized log-likelihood on 72
examples per model. K4V2 matches the BF16 average. K2V4 is
3.24~pp lower (95\% CI $[-9.72,0.00]$), and all of this gap comes from Qwen2.5;
Llama and Qwen3 saturate.

\paragraph{Long-context speculative decoding.}
Table~\ref{tab:app-long} uses PG-19 prefixes~\citep{rae2020compressive} with the Qwen2.5 3B/1.5B pair.
Prompts are consecutive token blocks from the streamed PG-19 training split. We
use three disjoint shards of 12 prompts at 16K and 8 prompts at 32K. At 16K
and 32K, K4V4 saves about 71\% of the draft cache and 31\% of total KV memory.
Acceptance drops by less than 1~pp; only the 16K interval excludes zero.

\begin{table}[ht]
  \centering
  \small
  \caption{K4V4 at long context. Acceptance changes are differences in pp
  from BF16 draft KV.}
  \label{tab:app-long}
  \begin{tabular}{rrrrr}
    \toprule
    Context & Prompts & Draft saved & Combined saved & $\Delta$ acceptance \\
    \midrule
    16K & 36 & 70.83\% & 30.99\% & $-0.94\ [-2.06,-0.13]$ \\
    32K & 24 & 70.96\% & 31.04\% & $-0.73\ [-2.99,1.28]$ \\
    \bottomrule
  \end{tabular}
\end{table}

\section{Memory Implementation}
\label{app:systems}

\paragraph{Packed storage.}
Figure~\ref{fig:memory-acceptance}b uses a real packed cache with
\texttt{uint8} payloads, BF16 metadata, and a BF16 residual window. The measured
tensors use 28 layers, two KV heads, and head dimension 128. This confirms that
the estimated saving we report is present in stored tensors rather than only in
theoretical bit counts. At 32,768 tokens, the 896.00 MiB BF16 cache becomes
260.20 MiB. Keys take 111.56 MiB of 4-bit payload, 27.89 MiB of BF16 scales and
zero-points, and 1.75 MiB for the BF16 residual window. Values take 112.00 MiB
of 4-bit payload and 7.00 MiB of BF16 scales and zero-points. This metadata costs
$2\cdot 16/32=1$ bit per quantized key element and $2\cdot 16/128=0.25$ bits
per value element, so a 4-bit payload does not give a 75\% saving. The packed cache
needs no padding or alignment bytes. We exclude temporary kernel workspace.

\paragraph{Kernel diagnostic.}
A Triton prototype reads the packed cache without first rebuilding a full BF16
cache. At 32K, it takes 0.708 ms per layer. Rebuilding the cache first takes
1.448 ms. Native BF16 SDPA takes only 0.103 ms. The prototype proves that packed
attention is possible, but it does not yet provide an end-to-end speedup.

\section{Longer Continuations and End-to-End Packed Prototype}
\label{app:followup}

\paragraph{Protocol.}
We also test 256-token continuations with the Qwen2.5 3B/1.5B pair, using
greedy decoding, a proposal length of four, and batch size one. Prompts come
from a seeded shuffle of the streamed FineWeb-Edu \texttt{sample-10BT} training
split, skipping 100 blocks for the quality study and 200 for the prototype. The quality study covers 1K context with
prompt seeds 0 and 1, and 16K context with seed 0, each with residual windows
of 0 and 128 tokens. Every setting tests BF16, K8V4, K4V8, and K4V4 on the same
24 prompts with fake quantization and the sequential verifier. The two residual
settings share prompts and BF16 baselines, so the six settings are not
independent. All use grouped per-channel keys.

\begin{table}[ht]
  \centering
  \small
  \caption{Acceptance in \% for 256-token continuations with 24 prompts per
  row. Differences are in pp with paired 95\% CIs. The last column
  compares K4V4 at residual windows of 128 and 0.}
  \label{tab:app-followup-quality}
  \resizebox{\linewidth}{!}{%
  \begin{tabular}{rrrrrrrrr}
    \toprule
    Context & Residual & Seed & BF16 & K8V4 & K4V8 & K4V4 & K4V4 $-$ BF16 & K4V4: residual 128 $-$ 0 \\
    \midrule
    1K & 0 & 0 & 76.69 & 77.25 & 77.01 & 76.61 & $-0.08\ [-1.18,+0.94]$ & \\
    1K & 0 & 1 & 74.01 & 74.18 & 73.58 & 73.45 & $-0.56\ [-1.30,+0.17]$ & \\
    1K & 128 & 0 & 76.69 & 77.44 & 76.61 & 77.02 & $+0.33\ [-0.43,+1.06]$ & $+0.408\ [-0.379,+1.290]$ \\
    1K & 128 & 1 & 74.01 & 73.79 & 73.55 & 73.42 & $-0.59\ [-1.34,+0.20]$ & $-0.033\ [-0.614,+0.629]$ \\
    16K & 0 & 0 & 79.62 & 79.54 & 79.64 & 79.20 & $-0.42\ [-1.42,+0.52]$ & \\
    16K & 128 & 0 & 79.62 & 79.92 & 79.42 & 79.55 & $-0.07\ [-1.04,+0.95]$ & $+0.351\ [-0.358,+1.110]$ \\
    \bottomrule
  \end{tabular}}
\end{table}

\paragraph{Acceptance and residual window.}
Table~\ref{tab:app-followup-quality} reports the mean of per-prompt acceptance.
This weights prompts equally, so it can differ from the pooled acceptance of
the cross-family study. The K4V4 change ranges from $-0.59$ to $+0.33$~pp, and every interval includes
zero. These exploratory intervals are not adjusted for multiple comparisons, so
we do not claim equivalence. The differences between residual windows of 128
and 0 tokens also include zero, so the window gives no clear acceptance gain.

\paragraph{Accepted prefixes and target calls.}
The accepted prefix is the number of draft tokens accepted before the first
rejection in a round, excluding the correction or bonus token. We average it
within each prompt and then across prompts. Across the six settings, BF16
accepts 2.95 to 3.17 tokens per round and K4V4 accepts 2.92 to 3.17. The paired
differences range from $-0.027$ to $+0.018$ tokens, and all six 95\% CIs
include zero. With K4V4, every proposed token is accepted in 62.8\% to 71.6\%
of rounds. The first rejection falls on the first, second, third, and fourth
proposed token in 10.8\% to 13.5\%, 7.6\% to 10.3\%, 5.7\% to 7.6\%, and 3.9\%
to 5.9\% of rounds. The sequential verifier makes 257 target calls for 256 tokens,
so these runs measure draft quality and correctness rather than target-call
savings.

\paragraph{End-to-end packed prototype.}
The prototype uses the packed cache and Triton kernel of Appendix~\ref{app:systems} with the
batched verifier, and packs key groups only after rollback.
Table~\ref{tab:app-followup-systems} compares target-only decoding, BF16
speculative decoding (SD), and packed K4V4 on 24 prompts at 1K and 12 prompts
at 16K. Latency is the mean wall-clock time per continuation, including prefill
but excluding warmup and the extra sequential audits. Savings compare stored
draft cache bytes with the BF16 draft cache, not with total GPU memory. Peak
memory is peak allocated GPU memory. We measure the target-only value before
loading the draft.

\begin{table}[ht]
  \centering
  \small
  \caption{End-to-end prototype with 256-token continuations at batch size one.
  $\Delta$ is the paired latency increase over BF16 SD with a 95\% CI. ``Exact''
  counts outputs that equal target-only greedy decoding.}
  \label{tab:app-followup-systems}
  \resizebox{\linewidth}{!}{%
  \begin{tabular}{llrrrrr}
    \toprule
    Context & Configuration & Latency (s) & $\Delta$ vs.\ BF16 SD (s) & Draft bytes saved & Peak allocated (GiB) & Exact \\
    \midrule
    1K & Target only & 7.10 & -- & -- & 5.98 & reference \\
    1K & BF16 SD & 9.68 & -- & 0.00\% & 8.87 & 3/24 \\
    1K & Packed K4V4, residual 0 & 12.77 & $+3.09\ [2.77,3.42]$ & 70.45\% & 8.87 & 4/24 \\
    1K & Packed K4V4, residual 128 & 12.87 & $+3.18\ [2.73,3.77]$ & 67.05\% & 8.87 & 5/24 \\
    \midrule
    16K & Target only & 8.65 & -- & -- & 7.69 & reference \\
    16K & BF16 SD & 14.54 & -- & 0.00\% & 10.76 & 2/12 \\
    16K & Packed K4V4, residual 0 & 17.20 & $+2.66\ [2.30,3.03]$ & 71.05\% & 10.76 & 2/12 \\
    16K & Packed K4V4, residual 128 & 17.46 & $+2.92\ [2.55,3.22]$ & 70.78\% & 10.76 & 2/12 \\
    \bottomrule
  \end{tabular}}
\end{table}

At 1K, packed K4V4 is 31.9\% and 32.9\% slower than BF16 SD with residual
windows of 0 and 128 tokens. At 16K, it is 18.3\% and 20.1\% slower. BF16 SD is
itself slower than target-only decoding, so the prototype
trades runtime for storage and does not give a speedup. Peak allocated memory
is the same as for BF16 SD. The prototype first builds the BF16 draft cache
during prefill and then packs it, so smaller stored tensors need not lower the
peak. We did not profile other contributors.

\paragraph{Correctness.}
With the sequential verifier, all 576 evaluations of the quality study (six
settings, four configurations, and 24 prompts) reproduce target-only greedy
decoding exactly. All 94 sequential audits recorded in the packed runs also
pass. The batched verifier behaves differently. At 1K, it matches target-only
greedy decoding on 3, 4, and 5 of 24 sequences for BF16, residual 0, and
residual 128. At 16K, it matches on 2 of 12 sequences for each configuration. A
mismatch means that at least one token differs. The BF16 baseline also
mismatches, so quantization alone cannot explain the difference. We have not
yet found the cause, and we do not assume that the mismatches are harmless near
ties, so we treat the timings as preliminary. The timings also come from a
shared node and one model pair. We do not measure production throughput,
maximum batch size, or concurrent serving.

\section{Choosing a Precision Policy}
\label{app:objectives}

We ask whether precision should be chosen using speculative acceptance or
ordinary draft quality. Over a grid of 20 uniform K/V settings of Qwen2.5 3B/1.5B on
WikiText-2, the two have Spearman correlation 0.964 at 1K and 0.878 at 4K. For five
pairs on FineWeb-Edu, equal-byte per-layer allocations built from each measure
coincide at mild budgets. At a three-bit budget, the acceptance-profiled choice is 0.48~pp worse
(95\% CI $[-0.98,-0.05]$) and has higher KL. Direct
acceptance is still the final serving metric, but we do not find a benefit from
an acceptance-only allocator.

\section{Compute and Asset Disclosure}
\label{app:disclosure}

\paragraph{Compute.}
GPU jobs used one or two NVIDIA RTX A5000 accelerators with 24 GiB of device
memory, 8 CPU cores, and 64 GiB of host memory. At most four GPUs ran
concurrently. SLURM accounting gives 229.45 total GPU-hours: 224.77 for
completed allocations and 4.68 for failed, cancelled, or timed-out
allocations. Table~\ref{tab:app-compute} separates the major paper-facing
experiment families. The remaining 85.55 GPU-hours cover robustness,
verification, proposal-length sweeps, and preliminary experiments. CPU-only
aggregation is excluded from GPU-hours. The recorded environment used Python
3.13.7, PyTorch 2.10.0 with CUDA 12.8, Transformers 5.3.0, and Datasets 4.8.2.

\begin{table}[ht]
  \centering
  \small
  \caption{Measured compute for major experiment families. GPU-hours equal
  wall time multiplied by allocated GPUs, including model loading. The
  experiments of Appendix~\ref{app:followup} are not included.}
  \label{tab:app-compute}
  \begin{tabular}{lrr}
    \toprule
    Experiment family & GPU allocations & GPU-hours \\
    \midrule
    Cross-family speculative and quality & 107 & 33.24 \\
    Key geometry and residual controls & 47 & 4.60 \\
    Passkey retrieval, layers, and policy & 95 & 13.14 \\
    LongBench retrieval & 18 & 2.51 \\
    Long-context speculative and quality & 19 & 1.90 \\
    Objective and precision matrix & 710 & 88.35 \\
    Packed-cache storage and kernels & 29 & 0.16 \\
    \bottomrule
  \end{tabular}
\end{table}

\paragraph{Third-party assets.}
Table~\ref{tab:app-assets} summarizes the upstream terms for every model family
and dataset used in reported results. We redistribute no model weights or other
dataset records, but the raw outputs released for Appendix~\ref{app:followup}
include the token IDs of their FineWeb-Edu prompts.

\begin{table}[ht]
  \centering
  \small
  \caption{Upstream licenses and terms for reported third-party assets.}
  \label{tab:app-assets}
  \begin{tabular}{lp{0.57\linewidth}}
    \toprule
    Asset & License or access terms \\
    \midrule
    Qwen2.5 1.5B, 7B & Apache License 2.0 \\
    Qwen2.5 7B-Instruct; Qwen3 4B, 8B & Apache License 2.0 \\
    Qwen2.5 3B & Qwen Research License \\
    Llama 3.1 8B(-Instruct); Llama 3.2 3B & Llama Community License,
      acceptable-use terms, and gated access \\
    OLMo 2 7B, 1B; SmolLM2 1.7B, 360M & Apache License 2.0 \\
    FineWeb-Edu & Open Data Commons Attribution License (ODC-By) v1.0 \\
    WikiText-2 & Creative Commons Attribution-ShareAlike License 3.0 \\
    PG-19 & Apache License 2.0 for the upstream benchmark compilation \\
    LongBench & MIT License \\
    Synthetic passkey & Generated by our released evaluator; no external records \\
    \bottomrule
  \end{tabular}
\end{table}
\FloatBarrier

\section{What We Do Not Claim}
\label{app:scope}

\begin{itemize}
  \item We do not claim that one key and value bit split works for every model.
  \item We do not claim faster kernels or decoding, or lower peak GPU memory.
  \item We do not claim that the 128-token residual window improves acceptance.
  \item We do not claim that our results hold under stochastic sampling.
  \item We do not quantize the target cache in the main method.
  \item We treat BF16 near ties separately from true target mismatches.
  \item We reject incomplete runs before computing aggregate results.
\end{itemize}

\end{document}